\documentclass{article}

\PassOptionsToPackage{numbers, compress}{natbib}

\usepackage[final]{neurips_2025}

\usepackage[utf8]{inputenc} 
\usepackage[T1]{fontenc}    
\usepackage{hyperref}       
\usepackage{url}            
\usepackage{booktabs}       
\usepackage{amsfonts}       
\usepackage{nicefrac}       
\usepackage{microtype}      
\usepackage{xcolor}         

\DeclareTextFontCommand{\emph}{\em}
\usepackage{float}
\usepackage{tikz-dependency}
\usepackage{amssymb}
\usepackage{placeins}
\usepackage{type1cm}
\usepackage{multirow}
\usepackage{balance}
\usepackage{subcaption}
\usepackage{caption}

\DeclareMathAlphabet{\mathpzc}{OT1}{pzc}{m}{it}
\usepackage{mathtools}
\usepackage{bm}
\usepackage{fancybox}
\usepackage{cleveref}
\usepackage{mathtools}
\usepackage{bold-extra}
\usepackage{wrapfig}
\usepackage{enumitem}
\usepackage{arydshln}
\usepackage{colortbl}
\usepackage{booktabs}
\usepackage{ascmac}
\usepackage{xspace}
\usepackage{longtable}
\usepackage{listings}

\PassOptionsToPackage{breaklinks}{hyperref}
\RequirePackage{hyperref}
\definecolor{darkblue}{rgb}{0, 0, 0.5}
\hypersetup{colorlinks=true, citecolor=darkblue, linkcolor=darkblue, urlcolor=darkblue}
\def\href#1#2{{#2}}
\renewcommand\cite{\citep}	

\title{A Multimodal Benchmark for Framing of Oil \& Gas Advertising and Potential Greenwashing Detection}

\author{%
Gaku Morio$^{1}$\thanks{This work was undertaken in part during the author’s time at Stanford University and Hitachi America, Ltd.} \quad Harri Rowlands$^{3}$\thanks{This work was undertaken in part during the author’s time at InfluenceMap CIC.} \quad Dominik Stammbach$^4$ \\
\textbf{
\quad Christopher D. Manning$^2$ \quad Peter Henderson$^4$
} \\
$^1$Hitachi, Ltd. \quad $^2$Stanford University \\
$^3$Centre for the Acceleration of Social Technology \quad $^4$Princeton University\\
\texttt{gaku.morio.vn@hitachi.com}\\
\texttt{manning@stanford.edu}\\
\texttt{harri@wearecast.org.uk}\\
\texttt{\{dominsta,peter.henderson\}@princeton.edu}
}

\makeatletter
\def\@trackname{}
\makeatother

\begin{document}

\maketitle

\begin{abstract}

Companies spend large amounts of money on public relations campaigns to project a positive brand image.
However, sometimes there is a mismatch between what they say and what they do. Oil \& gas companies, for example, are accused of ``greenwashing'' with imagery of climate-friendly initiatives.
Understanding the framing, and changes in framing, at scale can help better understand the goals and nature of public relations campaigns.
To address this, we introduce a benchmark dataset of expert-annotated video ads obtained from Facebook and YouTube.
The dataset provides annotations for 13 framing types for more than 50 companies or advocacy groups across 20 countries.
Our dataset is especially designed for the evaluation of vision-language models (VLMs), distinguishing it from past text-only framing datasets.
Baseline experiments show some promising results, while leaving room for improvement for future work: GPT-4.1 can detect environmental messages with 79\% F1 score, while our best model only achieves 46\% F1 score on identifying framing around green innovation.
We also identify challenges that VLMs must address, such as implicit framing, handling  videos of various lengths, or implicit cultural backgrounds.
Our dataset contributes to research in multimodal analysis of strategic communication in the energy sector.
\end{abstract}

\begin{wrapfigure}[15]{R}[0mm]{0.45\linewidth}
    \centering
    \includegraphics[width=\linewidth]{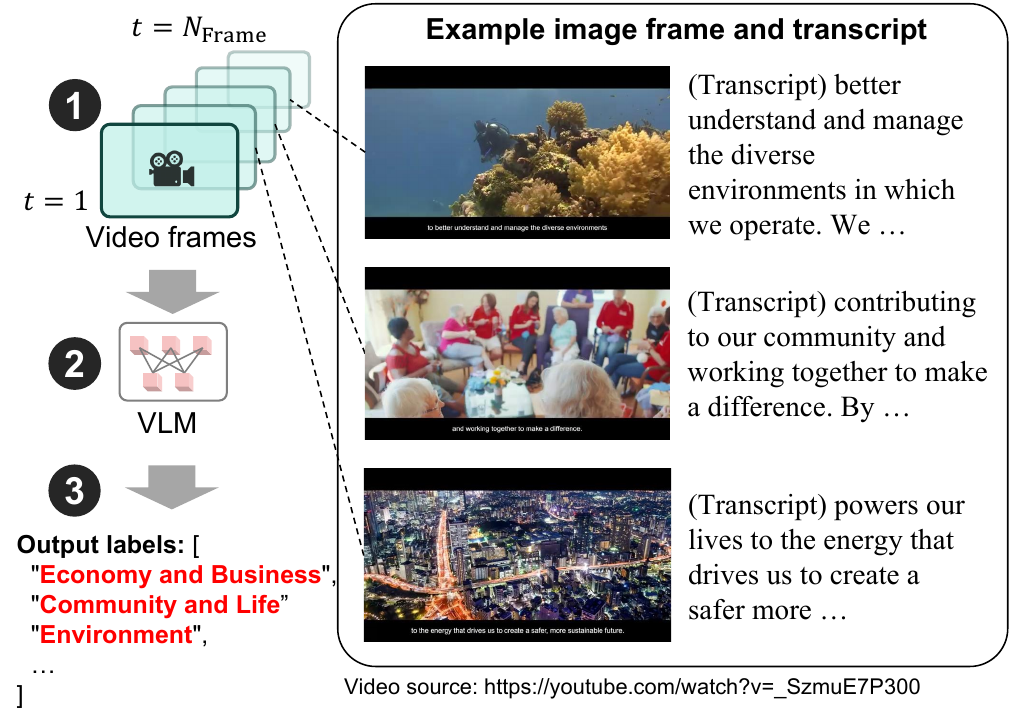} 
    \caption{Overview of the task of our dataset. } \label{fig:overview}
\end{wrapfigure}

\section{Introduction}

Framing plays an important role in how people perceive and evaluate information: \textit{``To frame is to select some aspects of a perceived reality [\ldots], to promote problem definition, causal interpretation, moral evaluation, and/or treatment recommendation [\ldots]''} \citep{entman_1993}. Recognizing this, oil and gas (O\&G) companies have been careful in messaging their values, visions, and how they operate \citep[see e.g.,][]{ ferns2019drilling, holder-etal-2023, supran2021rhetoric, gentile2025orchestrating}. Previous work qualitatively investigated self-portrayal and framing in O\&G campaigns and climate communication \citep{holder-etal-2023, si2023fossil}. 
For example, despite their dependence on fossil fuels, companies emphasize their contributions to reducing emissions or highlight that they are essential for keeping critical infrastructure such as hospitals running \cite{holder-etal-2023}. 
These messages are considered forms of \textit{greenwashing},\footnote{Usually defined as \textit{``behavior or activities that make people believe that a company is doing more to protect the environment than it really is''}\cite{greenwashing}.} as they tend to overstate corporate environmental contributions or distract from genuine climate action~\cite{de2020concepts,nemes-et-al-2022-integrated}.

Detection of greenwashing-related information has recently been transformed into computational benchmarks \citep{NEURIPS2023_7ccaa4f9}, e.g., a framing detection task from advertisement (ad) text \cite{rowlands-etal-2024-predicting}. 
These benchmarks can be used to assess and improve the capabilities of models to detect framing in public relations (PR) campaigns \citep{rowlands-etal-2024-predicting}. 
Accurate computational approaches have the potential to help interpret framing at scale \citep[e.g.,][]{piper-etal-2021-narrative, stammbach-etal-2022-heroes, gehring2023analyzing}. 
Interpreting the framing accurately has a significant potential role in social science and policymaking, especially for systematic understanding and monitoring of greenwashing.

However, previous related benchmarks do not consider non-text modalities, which is an important aspect of greenwashing \cite{holder-etal-2023,rowlands-etal-2024-predicting}. Previous qualitative analyses suggest the imagery of video ads can include strategic and misleading framings of greenwashing \cite{holder-etal-2023,muller-2025-people}. Without understanding video-based framing, it is impossible to holistically and systematically understand corporate strategic messaging or potential patterns of greenwashing -- especially if videos contain no spoken language (around 30\% of the videos in our dataset, see Table \ref{tb:basic_statistics}). 

Vision language models (VLMs) can be useful for detecting such video-based framing at scale \cite{vidllmsurvey}, but there exists no evaluation or benchmark for assessing this capability.
The lack of suitable benchmarking for video-based framing limits model development and potential social science applications of VLMs.

In this paper, we introduce a benchmark dataset for identifying framing techniques in O\&G advertising and marketing.
\Cref{fig:overview} illustrates the task setting.
The input is a video posted on social media by an O\&G entity, such as a company or advocacy group. The output is a set of framing techniques, e.g., `Community and Life' and `Environment'.
In our dataset, we cover major framing types for two sources: 
(i) The \textsc{Facebook} domain includes video ads published by O\&G companies and advocacy groups, where the videos are short and used in political campaigns to support the O\&G industry. These include seven greenwashing-related framing types for 17 entities, labeled by prior work \cite{holder-etal-2023,rowlands-etal-2024-predicting}.
(ii) The \textsc{YouTube} domain includes marketing videos from 42 O\&G related companies, where the videos are longer than those of \textsc{Facebook} and used to convey specific impressions to viewers. We annotate six different framing types such as `Environmental' or `Community and Life'.
By covering both domains, our dataset supports robust cross-domain evaluation of VLMs and provides a foundation for future research in energy and climate communication.

Our contributions can be summarized as follows:

\noindent
\textbf{(i) Multimodal dataset for framing of ads of O\&G entities:} We provide the first benchmark dataset of framing analysis in O\&G video ads.
The dataset includes fine-grained, expert-annotated labels capturing various framings across two domains. 
The dataset is designed to evaluate VLMs on real-world strategic framings and supports cross-domain, entity-level, and temporal analysis.
\\
\noindent \textbf{(ii) Benchmark evaluation of VLMs:} We benchmark the capabilities of 6 recent VLMs on our dataset by employing zero-shot and in-context learning settings. 
We also test a custom-built 1-shot prompting mechanism. Overall, while our pipeline improves performance significantly, we find that VLMs still face challenges, with room for improvement to improve accuracy and consistency across other labels (for example `Green Innovation') and cultural contexts.
\\
\noindent \textbf{(iii) Towards systematic greenwashing detection:} 
Our benchmark can potentially be interpreted as a benchmark for detecting greenwashing.
In \Cref{sec:pilot_study}, we discuss the relationship between greenwashing detection and our benchmark as well as a pilot study, including temporal trend analysis and company-level framing profiling for greenwashing risk assessments.

The \textbf{code} (\url{https://github.com/climate-nlp/multimodal-oil-gas-benchmark})
and
\textbf{data} (\url{https://huggingface.co/datasets/climate-nlp/multimodal-oil-gas-benchmark})
are available.

\section{Dataset Construction} \label{sec:dataset}

We design the dataset considering the following needs:
\\
\noindent
(i) \textbf{Cross-domain and context-aware benchmarking:} Framings can vary across platforms.
To evaluate VLMs across different contexts, we consider two domains, \textsc{Facebook} and \textsc{YouTube}.
\textsc{Facebook} enables us to capture framing related to climate obstruction\footnote{Campaigns to delay climate actions by companies and advocacy groups for example, \cite{holder-etal-2023}.} from energy companies and their value-chain entities within O\&G ad campaigns. On the other hand, the \textsc{YouTube} domain enables us to capture corporate strategies and implicit messages through official energy-related corporate channels. 
This dual domain design allows for the benchmarking of both generalization and domain-specific performance.
\\
\noindent
(ii) \textbf{Reliable annotations with diverse coverage:} 
We construct a dataset spanning 706 videos, 35k seconds of footage, 5.6k transcript segments (segments based on speech breaks extracted by Whisper-1 \cite{whisper1}), and 1.1k annotations -- a relatively large endeavor given the challenge of gathering expert-level annotations in this area.

In addition, our dataset spans more than 50 entities across 20 countries and includes videos published from 2010 to 2025, offering diversity in cultural context and changing framing strategies over time.
\\
\noindent
(iii) \textbf{Supporting practical downstream scenarios:} 
The dataset is designed not only for benchmarking but also to support downstream tasks such as potential automated greenwashing detection. We release the dataset with metadata including entity information, timestamps, and URLs, enabling integration into broader research.

\subsection{\textsc{Facebook}}

This subset is developed upon the previous work of Holder et al. \cite{holder-etal-2023} and Rowlands et al. \cite{rowlands-etal-2024-predicting}. 
The former \cite{holder-etal-2023} created a dataset to analyze Facebook ads from 2020 to 2021 in the United States, which allegedly contain messages of climate obstruction by O\&G entities. The latter \cite{rowlands-etal-2024-predicting} subsequently converted the dataset into a multi-label classification task. 
Because their work did not mainly consider videos attached to the ads, we collected 320 videos from the original ones \cite{holder-etal-2023,rowlands-etal-2024-predicting} and combined these videos with the annotated labels to create a new video dataset.

\noindent
\textbf{Nature of Domain -- Climate Obstruction Framing.} Previous literature \cite{holder-etal-2023,rowlands-etal-2024-predicting} points out O\&G companies and their agents greenwash through Facebook ads where they disseminate messages emphasizing the necessity and significance of fossil fuels, ultimately obstructing climate actions.
We refer to these messages as `climate obstruction framing' similar to existing work \cite{rowlands-etal-2024-predicting}.
Specifically, Holder et al. \cite{holder-etal-2023} decomposed the framing into four broad categories. `Community \& Resilience' emphasizes the positive impact of the O\&G industry on tax revenues and job creation. `Green Innovation and Climate Solutions' highlights emissions reduction targets or misleadingly describes O\&G as ``clean''. `Pragmatism' portrays O\&G as reliable and affordable, or essential raw materials for non-power-related goods such as hand sanitizer.
`Patriotic Energy Mix' stresses the importance of domestic O\&G production for energy independence and security.

Rowlands et al. \cite{rowlands-etal-2024-predicting} used more fine-grained labels originally defined by Holder et al. \cite{holder-etal-2023}, breaking down the four broader categories into seven subcategories. Because fine-grained labels are more informative, our study also adopts these labels. Below are the fine-grained label definitions (most descriptions are \textit{quoted} from the original works \cite{holder-etal-2023,rowlands-etal-2024-predicting}) used in our dataset:
\\
\noindent
\textbf{-- CA}: \textit{Helps national/local economies/communities, including through philanthropic efforts.} 
\\
\noindent
\textbf{-- CB}: \textit{Creates or sustains jobs.}
\\
\noindent
\textbf{-- GA}: \textit{Emissions reductions and transitioning the energy mix.}
\\
\noindent
\textbf{-- GC}: \textit{`Clean' gas as a climate solution.}
\\
\noindent
\textbf{-- PA}: \textit{Oil \& gas as energy sources are a pragmatic choice and critical for maintaining functioning or optimal power systems.}
\\
\noindent
\textbf{-- PB}: \textit{Oil \& gas are needed as raw materials for alternative (non-power-related) uses and manufactured goods.}
\\
\noindent
\textbf{-- SA}: \textit{The production of domestic O\&G reserves benefits the US, including through energy independence or energy leadership.}

\noindent
\textbf{Dataset Construction.} 
Our dataset builds on the above defined labels, while our contribution goes beyond repackaging:
We are the first to align videos with the text-derived labels in this domain.  
We obtained videos corresponding to each ad from the original dataset \cite{rowlands-etal-2024-predicting}. Ads without videos or ads that had already been removed were excluded. 
We transcribe the videos using a speech-to-text model (i.e., Whisper-1 \cite{whisper1}) to create the transcript text in the experiments.
Each video can have up to four labels \cite{rowlands-etal-2024-predicting}, making this a multi-label classification task. 
Previous literature \cite{holder-etal-2023,rowlands-etal-2024-predicting} discussed the inter-annotator agreement (IAA) of the original text-based dataset; please refer to those studies for further details. 
One limitation of our dataset is that the original annotations were mainly based on textual content, not videos, thus labels should be considered ``distant''. Despite this, our experiments show that models can still extract meaningful patterns. For further analysis of the distant labels, see \Cref{appendix:facebook_distant_label_aval}.
Finally, we randomly split the dataset into training and test sets (by 50:50).
We stored the dataset in a JSON Lines file (see \Cref{appendix:dataset_detail_facebook}).

\subsection{\textsc{YouTube}}

This subset consists of 386 annotated YouTube videos from major energy-related companies, capturing the ``impressions'' conveyed to viewers.

\noindent
\textbf{Nature of Domain -- Framing by Impressions.} YouTube has become an essential venue for companies targeting the general public with ads \cite{Febriyantoro01012020}. In our preliminary investigations, we found that energy companies strategically use YouTube to disseminate various implicit messages aimed at bolstering public perception of their industry. Initially, we attempted annotations using the same schema of \textsc{Facebook}, but achieving a reasonable IAA was challenging due to the implicit nature of the video content. Unlike Facebook ads, YouTube videos rarely explicitly mention job creation or energy independence. They presented visual imagery, such as wind turbines, to implicitly suggest environmental commitment, or footage of workers at oil facilities to foster trust.

After rounds of annotations and discussions, we decided not to limit annotating climate obstruction framing in our dataset. Instead, we develop the following label definitions to better capture these implicit impressions, while retaining Holder et al. \cite{holder-etal-2023}'s high-level idea: 
\\
\noindent
\textbf{-- Community and Life}:  Impressions of the company or O\&G contributing to daily life, community, culture, transportation, sports, and charitable efforts. E.g., footage of daily life or framings about oil usage for cooking.
\\
\noindent
\textbf{-- Economy and Business}:  Impressions of contributing to economic prosperity, business development, or tax revenue by the company or O\&G. E.g., testimony from local businesses.
\\
\noindent
\textbf{-- Work}: Impressions of job creation or reliable workplaces and employees. E.g., an image of smiling workers at an oil plant.
\\
\noindent
\textbf{-- Environment}: Impressions of reducing GHG emissions, supporting renewable energy, environmental responsibility, or ``clean'' O\&G usage. E.g., an image of renewable energy like solar panels and wind turbines.
\\
\noindent
\textbf{-- Green Innovation}: Impressions of innovation and visions for a green future, including the development of efficient new energy technologies. E.g., an image of a research lab, developing new climate solutions.

\noindent
\textbf{-- Patriotism}:  Impressions of contributing positively to the country, promoting national pride, and enhancing energy independence. E.g., a framing that emphasizes national brands such as ``we produce the US brand.''

We conducted multiple discussions as well as input from a non-profit community to provide the above typologies.
Each video can have multiple labels.
Labels may be inherently overlapping (e.g., local reforestation activities can imply both `Community and Life' and `Environment'). Annotators were asked to follow their intuition and annotate all relevant labels.

\noindent
\textbf{Dataset Construction.} We obtain a list of target entities by referencing the list of LobbyMap (\url{https://lobbymap.org/}), a platform that evaluates corporate engagement in climate policy.
We initially retrieve up to 30 videos per company by searching ads on each corporate official YouTube channel, resulting in a total of 720 videos (some channels did not have 30 videos available). 
We wanted to collect a wide range of advertising-like content from companies, and did not want to limit videos to specific product promotions or TV commercials. At the same time, videos that were clearly not advertising-like were excluded during the annotation process, ensuring the quality of the annotated dataset.
Next, we randomly sample 500 videos for annotation. Videos that are deleted or clearly not ads (e.g., earnings calls) are excluded, leaving a final set of 386 annotated videos.
Annotation guidelines are refined over multiple rounds, achieving a final Fleiss' Kappa \cite{fleiss-kappa} agreement score of 0.61.\footnote{On the other hand, we found that this varies depending on the calculation method (e.g., see \Cref{appendix:dataset_construction_detail_of_youtube} that also reports a 0.46 agreement score).}
Although this is not an almost perfect agreement, our labels are designed to reflect soft impressionistic framings, not hard factual typologies. This aligns with how greenwashing is typically found in real-world.
\Cref{appendix:dataset_construction_detail_of_youtube} describes details on the data collection and annotation.
Finally, we transcribe the videos using Whisper-1.
We randomly split the dataset into training and test sets (by 50:50).
The dataset is stored in JSON Lines format (see \Cref{appendix:dataset_detail_youtube}).

\begin{table}[t]
\minipage{0.4\textwidth}%
\centering
\footnotesize
\tabcolsep 2pt
\caption{Dataset statistics} \label{tb:basic_statistics}
\begin{tabular}{lrr}
\toprule
 &  \textsc{YouTube} & \textsc{Facebook} \\
\midrule
\# Videos & 386 & 320 \\
\# Transcript avail. & 377 & 119 \\
\# Transcript segm. & 4836 & 771 \\
\# Length (sec.) & 29545 & 5931 \\
\midrule
\# Entities & 42 & 17 \\
\# Countries & 20 & 1 \\
\midrule
Annotated by  & Humans & Distant \\
Labels & 802 & 381 \\
\bottomrule
\end{tabular}
\endminipage\hfill
\minipage{0.55\textwidth}%
  \includegraphics[width=\linewidth]{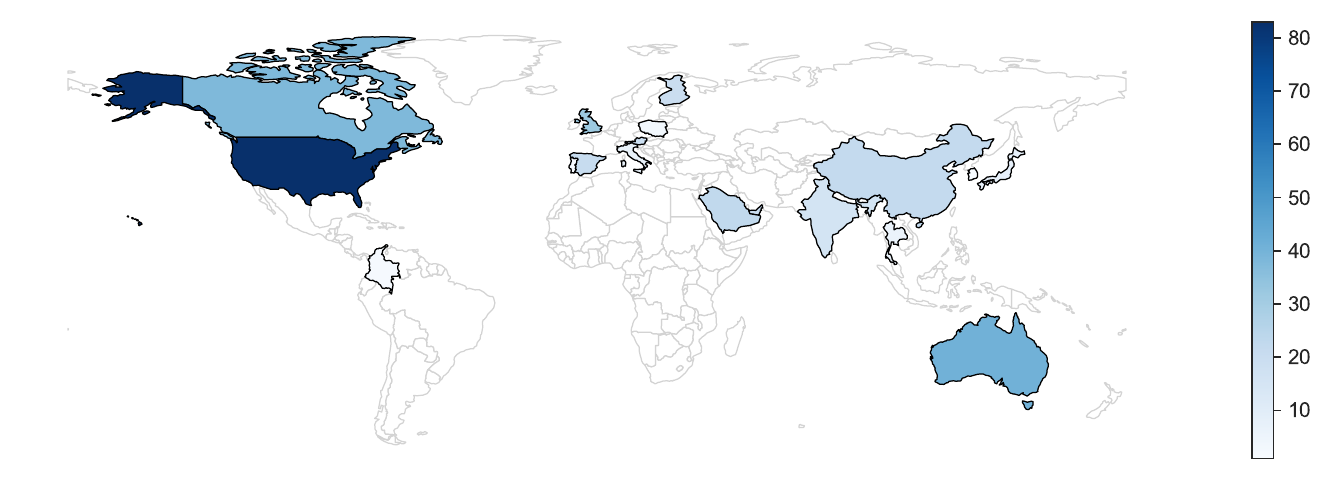}
  \captionof{figure}{The country distribution (based on headquarters location) of \textsc{YouTube}. Note that we primarily focus on English videos produced by multinational corporations.}\label{fig:heatmap_youtube_country}
\endminipage
\end{table}

\section{Dataset Analysis} \label{sec:dataset_analysis}


\Cref{tb:basic_statistics} shows the statistics for videos, entity-related information, and labels. We have a total of 706 videos totaling 35,476 seconds, 1,183 annotated labels, and 5,607 transcript segments.
Below we discuss the difficulty and challenges as a benchmark dataset:
\\
\noindent
\textbf{-- Modality Importance:} 
\Cref{tb:basic_statistics} shows that most videos of \textsc{YouTube} have transcripts, where around 37\% of videos do not contain transcripts in \textsc{Facebook}.
This might be because videos on \textsc{Facebook} are not the primary content of the ads, and text associated with the ad typically plays a more important role.
This property suggests that the importance of visual and transcript information could vary across domains. 
Moreover, the lack of transcripts highlights the need to incorporate the visual modality as well to improve model predictive performance. Handling both videos with transcripts and those without may present a new challenge for the model.
\\
\noindent
\textbf{-- Video Length:} \Cref{tb:basic_statistics} shows that \textsc{YouTube} videos tend to be longer than those of \textsc{Facebook} (the length distribution is shown in Appendix \Cref{fig:length_distribution}), suggesting that it could pose a challenge for VLMs, as they have to handle videos of varying lengths.
\\
\noindent
\textbf{-- Entity Coverage:} 
\Cref{tb:basic_statistics} shows that \textsc{YouTube} and \textsc{Facebook} datasets contain diverse entities.
All entity names are listed in \Cref{appendix:full_entity_list}.
Our dataset includes a wide range of energy-related companies and agents, ensuring that the benchmark does not overly depend on ads from a few dominant entities.
This presents another challenge as models must handle differences in advertising strategies across entities.
For example, companies that publicly disclose their investments in renewable energy will have different advertising strategies than those that do not.
\\
\noindent
\textbf{-- Geographic Coverage:} 
Some entities in the \textsc{Facebook} dataset were multinational companies. Since this domain targets ads served in the United States, we assume that the associated country for these ads is the United States only.
For \textsc{YouTube}, we rely on each entity's headquarters country, resulting in 20 countries in total as shown in \Cref{tb:basic_statistics}.
\Cref{fig:heatmap_youtube_country} shows the distribution of countries for \textsc{YouTube} videos, showing that most videos are from the United States, Canada, and Australia, while also covering other major economic powers such as China and India, and oil-producing countries in the Middle East.
Thus, the \textsc{YouTube} domain poses a challenge because VLMs must consider multiple cultural backgrounds.
On the other hand, note that most of the videos we cover are English-language content and may not reflect region-specific languages or cultural aspects.
\\
\noindent
\textbf{-- Imbalanced Labels:} 
In Appendix \Cref{tb:label_distribution}, we show the label distributions. The table indicates that `Green Innovation', `Patriotism', `PB', `GA', and `SA' are low-resource labels. Interestingly, patriotic messages (i.e., `Patriotism' and `SA') were rare in both domains.
This imbalanced property can be challenging for models, especially those relying on training data for in-context learning.
\\
\noindent
\textbf{-- Temporal Coverage:}
The \textsc{YouTube} domain covers videos from 2010 to 2025, highlighting the temporal diversity of the domain. (Please see Appendix \Cref{fig:yt_temporal_statistics} for the temporal label distribution.)

\section{Benchmark Experiments}

\subsection{Experimental Setup} \label{sec:experimental_setup}
We investigate how cutting-edge VLMs perform on our two-domain datasets.
Our task is a multi-label classification setting, where the input is a video and the output is a set of predicted labels (and the label set varies depending on the domain).
We use the F-score to evaluate classification performance.

Since the task requires nuanced interpretation of video content, if examples are shown to VLMs, it can significantly improve classification performance.
Therefore, we conduct two types of experiments:
\textbf{Zero-shot setting}, where no training examples are shown, and
\textbf{1-shot setting}, where one example from the training set is included in the prompt. Due to limited computational resources, we focus on the 1-shot setting, but also show K-shot results using a small model in \Cref{appendix:fewshot}.

\begin{figure*}[t]
    \centering
    \includegraphics[width=\linewidth]{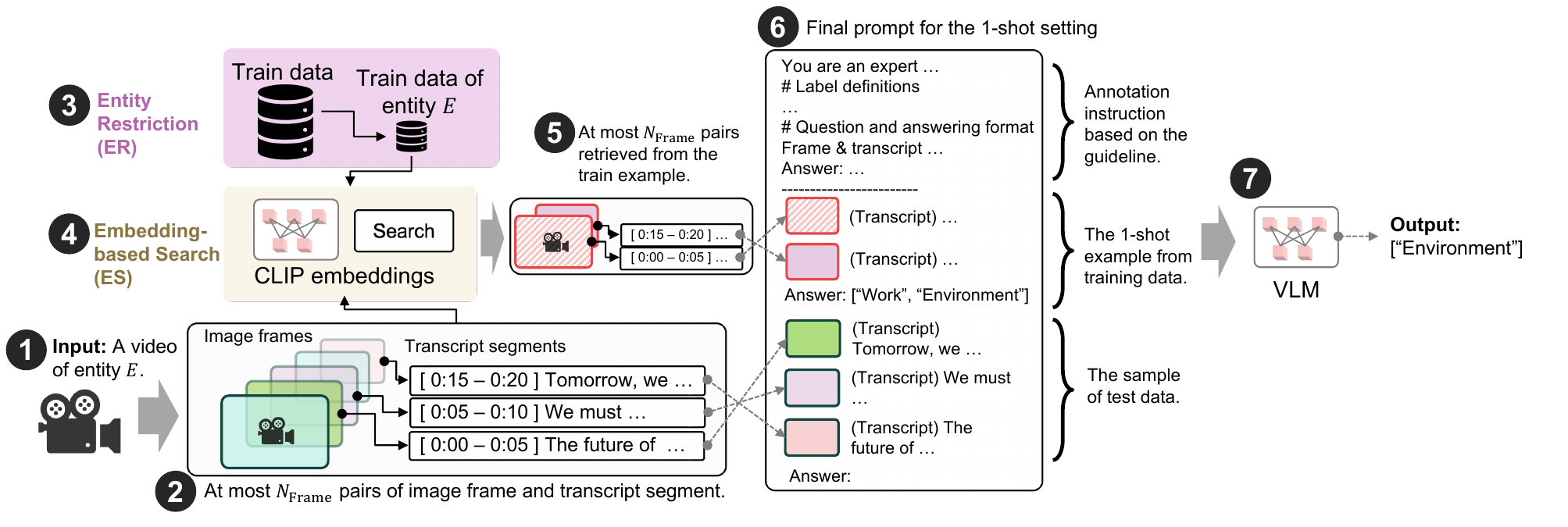}
    \caption{The overview of the entity-aware 1-shot prompt construction.} \label{fig:model}
\end{figure*}

\noindent
\textbf{Models.}
We benchmark various cutting-edge VLMs, from open-weight models to a closed model, as follows. 
\textbf{DeepSeek-VL2} \cite{deepseekvl2}: A relatively small (4.5B parameters) but high-performing multimodal model. 
\textbf{Qwen2.5-VL} \cite{Qwen2.5-VL}: A multimodal LLM based on Qwen2-VL \cite{Qwen2-VL} and Qwen-VL \cite{Qwen-VL}, designed for long video understanding, and known to outperform GPT-4o in some tasks. 
\textbf{InternVL2} \cite{internvl2}: An improved version of InternVL \cite{internvl}, trained with staged pretraining on large-scale vision-text datasets. 
\textbf{GPT-4.1} \cite{openai2025gpt4_1}: One of the latest LLMs with larger context windows. 
\textbf{GPT-4o-mini} \cite{openai2024gpt4omini}: A smaller variant of GPT-4o \cite{gpt4o}, which is a cost-efficient LLM that achieves near state-of-the-art performance on various multimodal tasks. 
For more details on the implementation, we refer to \Cref{appendix:implementation}.

\noindent
\textbf{Baseline Prompt Construction.}
Similar to existing literature \cite{zhang-etal-2024-omagent,hu-etal-2024-enhancing,shang-etal-2024-traveler,maaz-etal-2024-video}, we split each video into frame images, and then sample up to $N_{\text{Frame}}$  to input into the VLM.
At a high level, the prompt for the input consists of triplets: (i) Annotation instruction based on the guideline (c.f., Appendix \Cref{fig:prompt_fb} and \Cref{fig:prompt_yt}), (ii) up to $N_{\text{Frame}}$ sampled frame images, and, if available, the corresponding transcript segments, and (iii) (if 1-shot) a training example described in the same manner of (ii).
The frames are dynamically selected based on the transcript.
Specifically, for each transcript segment extracted by Whisper-1, we take the mean timestamp between its start and end points, and select the corresponding frame.
Due to computational resource constraints, we set $N_{\text{Frame}}$ differently depending on the model. For GPT-4.1, GPT-4o-mini, and Qwen2.5-VL, we set $N_{\text{Frame}}=10$. For InternVL2 and DeepSeek-VL2, we set $N_{\text{Frame}}=3$.
We also instruct the VLM to output the answer as a JSON list of labels.
An overview of the prompt construction and frame-transcript pairing can be found in part of \Cref{fig:model} (see (1), (2), (6), and (7) of the figure).

\noindent
\textbf{Entity-aware 1-shot Prompt Construction.}
As mentioned in \Cref{sec:dataset_analysis}, video characteristics can differ significantly across entities.
This will necessitate selecting an informative sample for 1-shot prompting as much as possible.
To verify this, we provide an entity-aware prompt construction approach for the 1-shot setting. The high-level idea of the approach is to select the 1-shot sample based on entity-aware similarity search from the training data, as is illustrated in \Cref{fig:model}.
The similarity search was inspired by a study that utilizes CLIP embeddings for retrieval-based prompting \cite{huang-etal-2024-towards} in an image and text classification task. 
We adapt the work into our video classification task by using both frame images and transcript segments for the embeddings.

Concretely, given a test video from entity $E$, our method proceeds as follows:
(i) \textbf{Entity Restriction (ER)} restricts the pool of candidate training videos to those belonging to the same entity $E$. This ensures that the examples used for in-context learning are relevant to the characteristics of the target entity.
(ii) \textbf{Embedding-based Search (ES)} performs a similarity search \cite{huang-etal-2024-towards} over the restricted training set using embeddings.
First, the frame images and transcript segments are embedded by CLIP: 
$\mathbf{e}_{\text{Frame}}=\frac{1}{N_\text{Frame}}\sum_{i=1}^{N_\text{Frame}} \mathbf{e}_{\text{Frame-}i}$ and $\mathbf{e}_{\text{Transcript}}=\frac{1}{N_\text{Frame}}\sum_{i=1}^{N_\text{Frame}} \mathbf{e}_{\text{Transcript-}i}$. 
Here, $\mathbf{e}_{\text{Frame-}i}$ and $\mathbf{e}_{\text{Transcript-}i}$ represents the $i$-th frame image embedding vector and transcript segment embedding vector, respectively. 
The video representation is then computed as $\mathbf{e} = \lambda_\text{Frame} \mathbf{e}_{\text{Frame}} + \lambda_\text{Transcript} \mathbf{e}_{\text{Transcript}}$. 
We set the hyperparameters $\lambda_\text{Frame}=0.5$ and $\lambda_\text{Transcript}=0.5$ through experiments.
Finally, the most similar training video is retrieved based on cosine similarity. The rest of the process for constructing the prompt is the same as the 1-shot process for the baseline prompt.
This approach makes VLMs see an example that is not only similar in content but also aligned with the features of the target entity.

\begin{table*}[t]
\centering
\footnotesize
\tabcolsep 1pt
\caption{Zero-shot and 1-shot experimental results in F-scores (\%). `All' denotes the micro-averaged score for all labels.} \label{tb:overall_results}
\begin{tabular}{ll|rrrrrrr|rrrrrrrr}
\toprule
    & & \multicolumn{7}{|c|}{\cellcolor[rgb]{0.93,0.85,0.85}\textbf{\textsc{YouTube}}} & \multicolumn{8}{|c}{\cellcolor[rgb]{0.85,0.85,0.93}\textbf{\textsc{Facebook}}} \\
\textbf{Model}  &  & \cellcolor[rgb]{0.93,0.93,0.93} All & \begin{tabular}{c}Comm.\\\&Life\end{tabular} & Work & Env. & \begin{tabular}{c}Green\\Innov.\end{tabular} & \begin{tabular}{c}Econ.\\\&Bus.\end{tabular} & \begin{tabular}{c}Patrio\\-tism\end{tabular} & \cellcolor[rgb]{0.93,0.93,0.93} All & CA & CB & PA & PB & GA & GC & SA \\
\midrule
\textbf{Zero-shot} & & \cellcolor[rgb]{0.93,0.93,0.93} &&&&&&& \cellcolor[rgb]{0.93,0.93,0.93}  \\
DeepSeekVL2 & \tiny 4.5B & \cellcolor[rgb]{0.93,0.93,0.93} 45.3 & 67.2 & 29.3 & 50.0 & 0.0 & 36.5 & 0.0 & \cellcolor[rgb]{0.93,0.93,0.93} 23.2 & 26.8 & 23.0 & 23.6 & 33.3 & 16.6 & 5.8 & 28.5 \\
InternVL2 & \tiny 8B & \cellcolor[rgb]{0.93,0.93,0.93} 53.1 & 71.7 & 55.2 & 50.0 & 35.2 & 39.7 & 27.2 & \cellcolor[rgb]{0.93,0.93,0.93} 22.5 & 31.2 & 6.4 & 8.6 & 4.8 & 37.6 & 11.4 & 31.8 \\
Qwen2.5-VL & \tiny 7B & \cellcolor[rgb]{0.93,0.93,0.93} 37.3 & 32.0 & 42.1 & 37.6 & 29.0 & 48.7 & 28.5 & \cellcolor[rgb]{0.93,0.93,0.93} 25.4 & 32.5 & 40.0 & 19.1 & 14.8 & 29.6 & 20.0 & 23.5 \\
Qwen2.5-VL & \tiny 32B & \cellcolor[rgb]{0.93,0.93,0.93} 60.7 & 68.8 & 61.7 & 73.5 & \textbf{46.9} & 47.3 & 42.8 & \cellcolor[rgb]{0.93,0.93,0.93} 49.0 & 42.8 & 78.6 & 35.0 & \textbf{60.0} & 46.6 & 38.0 & \textbf{51.4} \\
GPT-4o-mini & \tiny - & \cellcolor[rgb]{0.93,0.93,0.93} 60.5 & 72.4 & 66.2 & 72.8 & 39.2 & 41.1 & 43.1 & \cellcolor[rgb]{0.93,0.93,0.93} 54.2 & 40.9 & \textbf{79.3} & 56.5 & 40.0 & 42.5 & 52.1 & 43.9 \\
GPT-4.1 & \tiny - & \cellcolor[rgb]{0.93,0.93,0.93} \textbf{71.0} & \textbf{84.9} & \textbf{77.3} & \textbf{79.4} & 46.1 & \textbf{52.9} & \textbf{52.1} & \cellcolor[rgb]{0.93,0.93,0.93} \textbf{61.1} & \textbf{48.2} & 73.5 & \textbf{67.8} & 42.8 & \textbf{50.0} & \textbf{76.3} & 39.0 \\
\midrule
\textbf{1-shot} & & \cellcolor[rgb]{0.93,0.93,0.93}   \\
DeepSeekVL2 & \tiny 4.5B & \cellcolor[rgb]{0.93,0.93,0.93} 49.7 & 68.6 & 49.3 & 47.3 & 21.4 & 36.0 & 32.2 & \cellcolor[rgb]{0.93,0.93,0.93} 62.3 & 47.3 & 51.7 & 65.5 & \textbf{61.5} & 63.1 & 78.8 & 51.2 \\
InternVL2 & \tiny 8B & \cellcolor[rgb]{0.93,0.93,0.93} 56.7 & 78.1 & 61.9 & 53.0 & 34.1 & 39.3 & 25.0 & \cellcolor[rgb]{0.93,0.93,0.93} 46.2 & 30.0 & 29.2 & 60.5 & 35.2 & 41.2 & 71.6 & 35.0 \\
Qwen2.5-VL & \tiny 7B & \cellcolor[rgb]{0.93,0.93,0.93} 59.2 & 67.3 & 62.4 & 65.6 & \textbf{46.5} & 47.7 & 43.9 & \cellcolor[rgb]{0.93,0.93,0.93} 58.2 & 38.4 & 62.5 & 64.7 & 33.3 & \textbf{66.6} & 75.0 & 44.4 \\
Qwen2.5-VL & \tiny 32B & \cellcolor[rgb]{0.93,0.93,0.93} 66.2 & 76.0 & 70.2 & 77.4 & 45.8 & 48.9 & 48.2 & \cellcolor[rgb]{0.93,0.93,0.93} 70.5 & 44.8 & \textbf{85.2} & 67.2 & 54.5 & \textbf{66.6} & \textbf{90.3} & \textbf{64.5} \\
GPT-4o-mini & \tiny - & \cellcolor[rgb]{0.93,0.93,0.93} 63.0 & 72.9 & 68.6 & 74.9 & 39.9 & 45.2 & 51.2 & \cellcolor[rgb]{0.93,0.93,0.93} 65.2 & 53.3 & 80.6 & 67.2 & \textbf{61.5} & 52.3 & 74.0 & 51.2 \\
GPT-4.1 & \tiny - & \cellcolor[rgb]{0.93,0.93,0.93} \textbf{69.3} & \textbf{80.6} & \textbf{75.0} & \textbf{78.3} & 41.6 & \textbf{52.1} & \textbf{59.0} & \cellcolor[rgb]{0.93,0.93,0.93} \textbf{72.6} & \textbf{60.3} & 81.2 & \textbf{76.1} & 54.5 & 63.4 & 81.9 & 62.5 \\
\bottomrule
\end{tabular}
\end{table*}

\subsection{Results and Discussion}

\noindent
\textbf{Overall Results.}
\Cref{tb:overall_results} reports the experimental results on \textsc{YouTube} and \textsc{Facebook} domains.
Each table shows F-scores along with label-wise scores for both zero-shot and 1-shot settings.
`All' indicates the micro-averaged F-score across all labels.

In \textsc{YouTube}, GPT-4.1 performed best while GPT-4o-mini and Qwen2.5-VL (32B) achieved reasonably high F-scores in both zero-shot and 1-shot settings.
Given the nuanced nature of our annotation task and the IAA of 0.61, these results are promising.
We observed a large performance gap between 1-shot and zero-shot settings, especially for open-weight models such as Qwen2.5-VL.
This suggests that providing a suitable example in the prompt is important.
The exception is GPT-4.1 where the zero-shot setting outperformed the 1-shot setting. This demonstrates the remarkable capability of GPT-4.1 in this domain.
For the label-wise performance, labels such as `Community and Life', `Work', and `Environment' were classified with relatively higher accuracy. These labels often correlate with clear visual or textual cues, such as images of families, workers, or mentions of environmental commitments. In contrast, low-resource labels like `Patriotism', `Economy and Business', and `Green Innovation' may be difficult for models to classify. This could be because the messages rely more on subjective and subtle interpretation.

In \textsc{Facebook}, a similar trend was observed, while the 1-shot results consistently outperformed the zero-shot results across all models with a larger performance gap.
VLMs may struggle to classify the fine-grained labels, suggesting the importance of showing the 1-shot example in the prompt.
Interestingly, despite its smaller size, DeepSeekVL2 performed similarly to GPT-4o-mini in the 1-shot setting.
This suggests that even smaller-scale VLMs can be effective under certain conditions.
The label-wise F-scores show that labels such as `PB' showed lower F-scores across models, suggesting that these categories are particularly difficult for the models to interpret.

Below, we further discuss ablation studies and error analyses. Otherwise specified, we use the 1-shot prompting with ES and ER.

\begin{wraptable}[14]{R}[0mm]{0.48\linewidth}
\centering
\footnotesize
\tabcolsep 1pt
\caption{Ablation results in the 1-shot prediction. \textbf{T} represents the transcript input, \textbf{ES} represents the embedding search in the 1-shot sample selection, and \textbf{ER} represents the entity restriction in the 1-shot sample selection. The full results for all models can be found in \Cref{tb:ablation_full}.} \label{tb:ablation}
\begin{tabular}{lllllrr}
\toprule

\textbf{Model} &  & \textbf{T}  &  \textbf{ES} & \textbf{ER}  & \textsc{YouTube} & \textsc{Facebook} \\
\rowcolor[rgb]{0.93,0.93,0.93}Qwen2.5-VL & 32B & $\checkmark$ & $\checkmark$ & $\checkmark$ & \textbf{66.2} & \textbf{70.5} \\
 &  & $\times$ & $\checkmark$ & $\checkmark$ & 61.2 & 60.6 \\
 &  & $\checkmark$ & $\times$ & $\checkmark$ & 64.0 & 59.1 \\
 &  & $\checkmark$ & $\checkmark$ & $\times$ & 65.6 & 68.1 \\
\bottomrule
\end{tabular}
\end{wraptable}

\noindent
\textbf{Ablation -- Modality Importance.}
\Cref{tb:ablation} shows the results of ablation study using Qwen2.5-VL 32B. The `$\text{\textbf{T}}=\times$' (i.e., the 2nd row of the table) means that the transcript input is ablated. In \textsc{YouTube}, we obtained 61.2\% F-score, while in \textsc{Facebook} we obtained 60.6\%, suggesting the modality from the transcript of the video is important for the inference of VLMs. There is a huge gap between models with and without transcript input, especially in \textsc{Facebook}. 
On the other hand, smaller models (e.g., DeepSeekVL2) on \textsc{YouTube}, including transcript inputs sometimes degrades performance (see \Cref{tb:ablation_full}).
These results above align with our idea in \Cref{sec:dataset_analysis} that the importance of visual and transcript information varies across domains.

\noindent
\textbf{Ablation -- Entity Restriction and Embedding-based Search.}
\Cref{tb:ablation} also shows the effect of the entity-aware 1-shot prompt construction (see \textbf{ES} and \textbf{ER} in the table).
Incorporating both components (ES, ER) generally led to better performance of Qwen2.5-VL 32B predictions, highlighting the importance of retrieving similar examples for the 1-shot prompt.
However, the effect of ER was mixed when observing all models (see Appendix \Cref{tb:ablation_full}).
When only a few training videos per entity are available, restricting the search space may limit the quality of the retrieved examples.

\begin{figure}[t]
\minipage{0.3\textwidth}%
  \includegraphics[width=\linewidth]{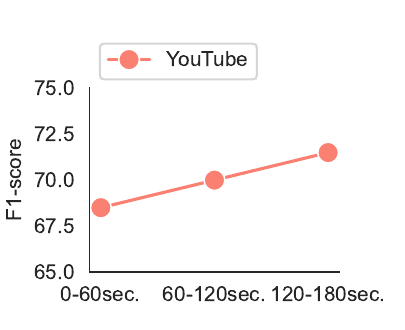}
    \caption{The video length and F-score of GPT-4.1.} \label{fig:score_by_video_len}
\endminipage\hfill
\minipage{0.3\textwidth}%
  \includegraphics[width=\linewidth]{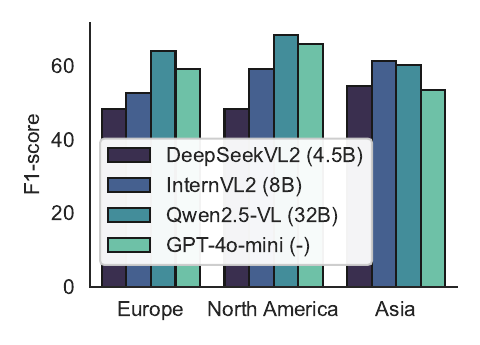}
    \caption{The region and F-score for \textsc{YouTube}. We exclude Middle East from Asia.} \label{fig:yt_region_score}
\endminipage\hfill
\minipage{0.36\textwidth}%
  \includegraphics[width=\linewidth]{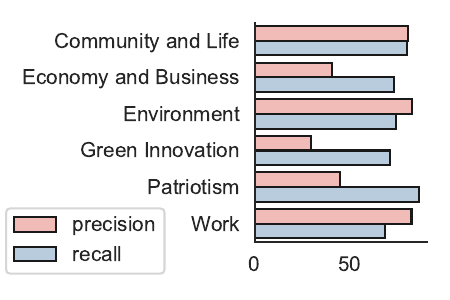}
  \caption{The precision and recall analysis for GPT-4.1 in \textsc{YouTube}.
  }\label{fig:yt_precision_recall}
\endminipage
\end{figure}

\noindent
\textbf{Error Analysis -- Video Length.}
\Cref{fig:score_by_video_len} shows the F-scores for different video lengths. Interestingly, shorter videos (i.e., 0--60 sec.) seem to be more challenging for the model. This might be because shorter videos are more contextualized and vague, making it difficult for models to predict framings.
The result suggests shorter videos need careful handling.

\noindent
\textbf{Error Analysis -- Geographic Effect on Models.}
\Cref{fig:yt_region_score} shows the F-scores in each geographical region (obtained from the countries where the headquarters are located) for \textsc{YouTube}. The figure shows GPT-4o-mini and Qwen2.5-VL perform well on videos from European and North American companies, while DeepSeekVL2 and InternVL2 outperform GPT-4o-mini for videos from Asia.
The result suggests that each VLM has a different ability to handle specific cultural contexts.
This insight supports our idea in \Cref{sec:dataset_analysis} that VLMs must consider multiple cultural backgrounds.

\noindent
\textbf{Error Analysis -- Over-label or Under-label.}
\Cref{fig:yt_precision_recall} shows a precision and recall analysis across different labels for \textsc{YouTube} to investigate the extent to which a VLM  over- or under-labels.
For labels with relatively higher overall F-scores, such as `Community and Life', `Environment', and `Work', precision tends to be higher than recall. For more difficult labels such as `Economy and Business', `Green Innovation', and `Patriotism', recall tends to be higher than precision.
This suggests that for difficult labels, VLMs tend to over-label, while experts tend to under-label when assigning such labels.

We also note significant variation in the labelling habits of different models. We investigated the number of output labels by the zero-shot prompting. We found Qwen2.5-VL 7B is by far the most under-labeler (171 labeled, mean: 349) while GPT-4o-mini is an over-labeling model (409 labeled). On the other hand, models struggled to capture the most common co-occurrences demonstrated in the gold dataset. 
DeepSeekVL2 shows a strong pairing of the labels `Community and Life' and `Economy and Business' (115 times vs. 63 times in the gold), despite this being rare in the gold. Meanwhile, DeepSeekVL2 does not accurately reflect the most common pair in the gold dataset -- `Community and Life' and `Work' (DeepSeekVL2: 17 vs gold: 131).

\noindent
\textbf{Takeaway.}
The above discussions identify challenges that VLMs must address, such as handling different video lengths, over-label or under-label annotations, and cultural backgrounds.
Through benchmark evaluations using our dataset, researchers can select the VLM that best suits their needs, recognize the limitations of each model, and test new methods.

\begin{figure}[t]
\minipage{0.45\textwidth}%
    \includegraphics[width=\linewidth]{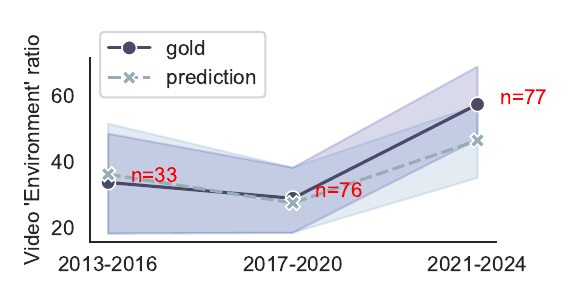}
    \caption{The ratio of videos with `Environment' on \textsc{YouTube} by  GPT-4.1. We show the 95\% confidence interval with bootstrap resampling.} \label{fig:yt_env_ratio_yrs}
\endminipage\hfill
\minipage{0.5\textwidth}%
    \includegraphics[width=\linewidth]{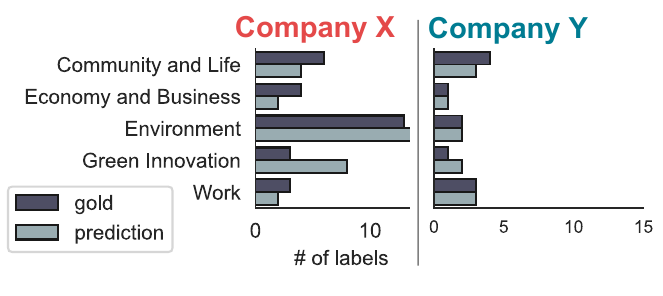}
    \caption{The GPT-4.1 predicted and gold labels for two example companies.} \label{fig:case_study}
\endminipage
\end{figure}

\section{Discussion: Towards Multimodal Greenwashing Detection} \label{sec:pilot_study}

Videos in our benchmark are potentially candidates for greenwashing. The framing of \textsc{Facebook} videos in our dataset represents the following forms of greenwashing: Influencing the general public to perceive O\&G as ``clean'' (i.e., `GC') or emphasizing the necessity of oil enhances welfare for communities or is a pragmatic necessity (i.e., `CA', `CB', `PA' and `PB'). This emphasizes the necessity of the O\&G industry and reduces awareness of environmental impact, which is related to obstruction and greenwashing by selective disclosure \cite{de2020concepts,nemes-et-al-2022-integrated}. Our benchmark allows evaluating whether VLMs can accurately detect such forms of greenwashing.

The framing of `Environment' and `Green Innovation' in \textsc{YouTube} often leverages vagueness to portray an environmentally friendly image without making specific commitments \cite{nemes-et-al-2022-integrated, stammbach-etal-2023-environmental}.
Analogously, the labels `Community and Life', `Economy and Business', and `Work' relate to obstruction and selective disclosure similarly to the \textsc{Facebook} labels `CA', `CB', `PA' and `PB'.

Furthermore, we introduce the following pilot studies that assess greenwashing risks through a comprehensive analysis:
\\
\noindent
\textbf{Temporal Trend Analysis.}
Understanding how environmental framing has shifted over time can provide valuable insights in O\&G communication strategies. To this extent, we analyze the temporal trend of environment-related messaging in Figure \ref{fig:yt_env_ratio_yrs}.
We calculate the environment label ratio, defined as the number of videos labeled as `Environment' divided by the total number of videos, for each year.
We observe that the predicted trend closely follows the ground-truth trend, suggesting that our model can reasonably capture temporal trends in framing.
Importantly, the model captures the increasing trend after 2020.
This correlates to the period of the Biden administration (although we do not verify its causality).  
Given that greenwashing typically involves emphasizing positive environmental communication \cite{de2020concepts}, this increase could be interpreted as a potential signal of industry-wide greenwashing.
\\
\noindent
\textbf{Company-level Analysis.}
Here, we present two case studies based on selected companies.
\Cref{fig:case_study} shows the gold and predicted label distributions for Company X and Company Y (anonymized), suggesting that the model can reasonably replicate the gold distribution.
Interestingly, Company X exhibits a high proportion of `Environment' labels, showing that the company may actively use its YouTube channel to promote a strong environmental image that exceeds the industry average. 
In contrast, Company Y places more emphasis on `Community and Life', reflecting a perspective that presents O\&G as essential to our daily life. Also, we found this company frequently includes other labels such as `Work' and `Environment' in the same videos.
We observed that in such multi-labeled videos, environment-related content was often vague and embedded as part of the overall impression, rather than linked to concrete environmental performance.

In our view, the identified cases above are promising candidates to further evaluate manually. Gathering such insights through computational assistance can reduce the amount of manual research required to reach such conclusions. Further, they show the practical potential of our benchmark dataset in greenwashing detection, corporate strategy profiling, and social science research.

\section{Conclusion}
We propose a video classification task for framing in ad videos by O\&G entities, and we introduce a benchmark dataset that consists of two sources, namely \textsc{YouTube} and \textsc{Facebook}. The dataset is challenging for VLMs and we see lots of room for improvement going forward. 

Furthermore, our work enables practical applications such as computer-assisted multimodal greenwashing analysis or social science work about temporal trends in O\&G messaging.

We hope this dataset encourages researchers and practitioners to study corporate messaging from both visual and textual perspectives.
Future directions include extending the task to more granular labels, e.g., whether the messaging is implicit or explicit or whether the message is portrayed visually, via spoken language and/or as captions. 
Lastly, we plan to include more domains with high greenwashing risks.

\begin{ack}
We would like to thank the anonymous referees for their helpful comments on this paper. 
GM conducted this research as part of the Stanford Data Applications Affiliates Program.
At the time the research was conducted, GM was an employee of Hitachi America, Ltd., which provided financial support for the study.
\end{ack}

{\small
\bibliographystyle{acl_natbib}
\bibliography{custom, anthology}
}


\newpage
\section*{NeurIPS Paper Checklist}

\begin{enumerate}

\item {\bf Claims}
    \item[] Question: Do the main claims made in the abstract and introduction accurately reflect the paper's contributions and scope?
    \item[] Answer: \answerYes{} 
    \item[] Justification: Our contributions are the creation of datasets, analysis, benchmark experiments, and discussions, which are appropriately documented in the main text.

\item {\bf Limitations}
    \item[] Question: Does the paper discuss the limitations of the work performed by the authors?
    \item[] Answer: \answerYes{} 
    \item[] Justification: \Cref{appendix:limitation}.

\item {\bf Theory assumptions and proofs}
    \item[] Question: For each theoretical result, does the paper provide the full set of assumptions and a complete (and correct) proof?
    \item[] Answer: \answerNA{} 
    \item[] Justification: This paper does not contain theoretical results.
    
\item {\bf Experimental result reproducibility}
    \item[] Question: Does the paper fully disclose all the information needed to reproduce the main experimental results of the paper to the extent that it affects the main claims and/or conclusions of the paper (regardless of whether the code and data are provided or not)?
    \item[] Answer: \answerYes{}{} 
    \item[] Justification: Major implementation details and hyperparameters are described in \Cref{sec:experimental_setup} and  \Cref{appendix:implementation}.

\item {\bf Open access to data and code}
    \item[] Question: Does the paper provide open access to the data and code, with sufficient instructions to faithfully reproduce the main experimental results, as described in supplemental material?
    \item[] Answer: \answerYes{} 
    \item[] Justification: We made our dataset and code public.

\item {\bf Experimental setting/details}
    \item[] Question: Does the paper specify all the training and test details (e.g., data splits, hyperparameters, how they were chosen, type of optimizer, etc.) necessary to understand the results?
    \item[] Answer: \answerYes{} 
    \item[] Justification: See \Cref{sec:experimental_setup} and \Cref{appendix:implementation}.

\item {\bf Experiment statistical significance}
    \item[] Question: Does the paper report error bars suitably and correctly defined or other appropriate information about the statistical significance of the experiments?
    \item[] Answer: \answerYes{} 
    \item[] Justification: In \Cref{fig:yt_env_ratio_yrs}, we report statistics suitably.

\item {\bf Experiments compute resources}
    \item[] Question: For each experiment, does the paper provide sufficient information on the computer resources (type of compute workers, memory, time of execution) needed to reproduce the experiments?
    \item[] Answer: \answerYes{} 
    \item[] Justification: \Cref{appendix:computational_resource}.
    
\item {\bf Code of ethics}
    \item[] Question: Does the research conducted in the paper conform, in every respect, with the NeurIPS Code of Ethics \url{https://neurips.cc/public/EthicsGuidelines}?
    \item[] Answer: \answerYes{} 
    \item[] Justification: \Cref{appendix:ethics_statement} and \Cref{appendix:dataset_documentation}.

\item {\bf Broader impacts}
    \item[] Question: Does the paper discuss both potential positive societal impacts and negative societal impacts of the work performed?
    \item[] Answer: \answerYes{} 
    \item[] Justification: \Cref{appendix:ethics_statement}.

\item {\bf Safeguards}
    \item[] Question: Does the paper describe safeguards that have been put in place for responsible release of data or models that have a high risk for misuse (e.g., pretrained language models, image generators, or scraped datasets)?
    \item[] Answer: \answerNA{} 
    \item[] Justification: Risk of misuse is low.

\item {\bf Licenses for existing assets}
    \item[] Question: Are the creators or original owners of assets (e.g., code, data, models), used in the paper, properly credited and are the license and terms of use explicitly mentioned and properly respected?
    \item[] Answer: \answerYes{} 
    \item[] Justification: \Cref{appendix:dataset_documentation}, \Cref{appendix:implementation}, and \Cref{sec:experimental_setup}.

\item {\bf New assets}
    \item[] Question: Are new assets introduced in the paper well documented and is the documentation provided alongside the assets?
    \item[] Answer: \answerYes{} 
    \item[] Justification: \Cref{appendix:dataset_documentation}.

\item {\bf Crowdsourcing and research with human subjects}
    \item[] Question: For crowdsourcing experiments and research with human subjects, does the paper include the full text of instructions given to participants and screenshots, if applicable, as well as details about compensation (if any)? 
    \item[] Answer: \answerNA{} 
    \item[] Justification: The paper does not involve crowdsourcing nor research with human subjects.

\item {\bf Institutional review board (IRB) approvals or equivalent for research with human subjects}
    \item[] Question: Does the paper describe potential risks incurred by study participants, whether such risks were disclosed to the subjects, and whether Institutional Review Board (IRB) approvals (or an equivalent approval/review based on the requirements of your country or institution) were obtained?
    \item[] Answer: \answerNA{} 
    \item[] Justification: The paper does not involve crowdsourcing nor research with human subjects.

\item {\bf Declaration of LLM usage}
    \item[] Question: Does the paper describe the usage of LLMs if it is an important, original, or non-standard component of the core methods in this research? Note that if the LLM is used only for writing, editing, or formatting purposes and does not impact the core methodology, scientific rigorousness, or originality of the research, declaration is not required.
    \item[] Answer: \answerNA{} 
    \item[] Justification:  The core method development in this research does not involve LLMs as any important, original, or non-standard components.  We used an LLM for writing and editing purposes. Though this is not required to disclose, we did in \Cref{appendix:declaration_of_LLM_usage}.

\end{enumerate}

\newpage
\appendix
\section{Appendix}

\subsection{Ethics Statement} \label{appendix:ethics_statement}

In alignment with the Code of Ethics guidelines \footnote{\url{https://neurips.cc/public/EthicsGuidelines}}, the following discuss ethical considerations and potential societal impact of this work.

\textbf{Privacy.} Videos in our dataset are voluntarily made publicly available by its producers as marketing and promotional videos. Thus, we do not see any violation of privacy issues.

\textbf{Consent.} The videos used in our study are publicly available and of public interest. Given the public interest about framing strategies for O\&G video, it is fair use to annotate and release such videos for research purposes and do not require consent. 

\textbf{Deprecated Datasets.} Not applicable.

\textbf{Copyright and Fair Use.} Given the public interest about framing strategies for O\&G video, it is fair use to annotate and release such videos for research purposes. 

\textbf{Representative Evaluation Practice.} We discuss the geographic diversity and temporal coverage of our dataset in our paper, and showcase the companies included (\Cref{tb:entity_facebook} and \Cref{tb:entity_youtube}) and their headquarters’ location in Figure \ref{fig:heatmap_youtube_country}. Our dataset, although limited in size, exhibits extensive coverage of large O\&G companies in various countries across four continents and 15 years. Nevertheless, there exists a bias towards promotional videos of the largest O\&G companies and English videos. In future work, we are excited to explore videos from smaller companies, short video formats (e.g., Facebook Reels, TikTok, YouTube Shorts) and non-English videos.

\textbf{Safety.} We do not see any major safety concerns in releasing annotations for promotional videos of O\&G videos for educational and research purposes.

\textbf{Security.} We do not see any major security concerns in releasing annotations for promotional videos of O\&G videos for educational and research purposes.

\textbf{Discrimination.} We do not see any major discrimination concerns in releasing annotations for promotional videos of O\&G videos for educational and research purposes.

\textbf{Surveillance.} We do not see any major surveillance concerns in releasing annotations for promotional videos of O\&G videos for educational and research purposes.

\textbf{Deception \& Harassment.} We do not see any major deception \& harrasment concerns in releasing annotations for promotional videos of O\&G videos for educational and research purposes.

\textbf{Environment.} We recognize the environmental impact, e.g., energy and/or water consumption while benchmarking various models on the released dataset.

\textbf{Human Rights.} Not applicable.

\textbf{Bias and fairness.} Due to annotation artifacts and the nature of distributions of large O\&G companies, our dataset contains biases related to countries and regions. Researchers employing the dataset should be aware of such biases.

\subsection{Limitations} \label{appendix:limitation}

The data acquisition process includes implicit or unrecognized biases. For example, the subset from \textsc{Facebook} was obtained dataset of the previous literature. The data collection process for the subset of \textsc{YouTube} includes the channel search to access to relevant videos, which may not reproduce depending on the internal recommendation algorithm of the platform. We do not collect all videos but sampled from the available video set. We do not consider videos which were removed or unable to download. 

Our dataset was sourced from Facebook and YouTube, a decision driven by the accessibility for data acquisition. Thus, our findings may not generalize to the broader video-based ads across diverse platforms. This study does not address the domain of short-form video content (e.g., Facebook Reels, TikTok, YouTube Shorts), despite its significant and increasing prevalence. Furthermore, our dataset can contain a bias towards content from large, multinational corporations and advertisers in economically dominant regions or oil producing countries. Future research would benefit from incorporating more diverse domains that include emerging markets, smaller firms, and underrepresented video formats.

Our dataset has been created with best efforts in terms of scale and annotation accuracy. However, model bias may arise due to inaccurate annotations or small sample sizes. Additionally, due to the nature of advertising videos, highly similar content may be included in both the training and test data, which could potentially lead to optimistic benchmark results. Transcripts are automatically generated and may contain errors in speech recognition. Furthermore, the dataset was created primarily with videos that are understandable to English or Japanese speakers, so linguistic representativeness is limited.
\textsc{Facebook} videos are originally labeled mainly based on ad text rather than the video's overall content. This ``distant annotation'' approach could ignore visual and nuanced cues within the video, potentially introducing bias into model training and evaluation.

Despite spending days debating interpretations and resolving disagreements for the \textsc{YouTube} domain, we can encounter subjective interpretations when labeling.
Certain framing categories (e.g., `Green innovation' and `Environment') may be interpreted as overlapping in some cases. This kind of potential ambiguity may introduce additional uncertainty into model development and evaluation.

Given the size of our dataset, the generalizability of our analyses and experimental results can be restricted. Given above, our benchmark is unlikely to be suitable for fine-tuning models on the task. However, we note that there is limited availability of datasets for multi-modal analysis more broadly across all climate change domains. Even though it is not directly related to our work, MultiClimate \cite{wang-etal-2024-multiclimate}, for example, is a recent dataset on climate stance detection from videos, containing a total of 100 videos.
More importantly, our work is aimed at covering a diverse set of videos rather than a large number of videos. Our work is characterized by fine-grained labels across two different domains. Moreover, within each domain, we consider a variety of entities.

From a technical standpoint, our study is subject to several limitations. Computational resource constraints restricted the scale of our frame-level analysis (c.f., the number of sampled image frames input into the VLMs), potentially limiting our ability to discern highly granular temporal patterns within video content. Secondly, we only tried the single run for each VLMs. There can be statistical fluctuations for the results.

\subsection{Code and Data Availability} \label{appendix:code_and_data}
The code and data are available on Github nd Huggingface.

\subsection{Related Work} \label{appendix:related_work}

Recent progress in capabilities to analyze videos in open and closed source models make this work possible. Among others, these include GPT-4o-mini \citep{openai2024gpt4omini}, DeepSeek-VL2,  \cite{deepseekvl2}, InternVL2 \citep{internvl2} and the Qwen2.5-VL model family \citep{Qwen2.5-VL} and we benchmark all of these in our work. Many of the more recently released next-generation language models are capable of processing videos, e.g., Claude and Gemini and Qwen3 \citep{anthropic2024claude3, google2025gemini25, Qwen2.5-VL} and we expect this trend to continue and models to become better at analyzing videos in the near future. We leave it to future work to examine such models on our benchmark more carefully.

Further, our work connects to narratives and framing in social sciences and economics \citep{blair2016applying, holder-etal-2023, enfield2024language} and to work on computational narrative analysis \citep{piper-etal-2021-narrative, stammbach-etal-2022-heroes}, framing \citep{card-etal-2015-media, arora2025multimodalframinganalysisnews} and agenda setting \citep{tsur-etal-2015-frame}. Specifically in the environmental context, we find work facilitating the automatic analysis of company reports to extract quantities of interest, such as environmental claims \citep{stammbach-etal-2023-environmental}, environmental narratives \citep{gehring2023analyzing, rowlands-etal-2024-predicting} and corporate climate policy engagement  \citep{NEURIPS2023_7ccaa4f9}. All these studies hint at the feasibility of computational assistance for undertaking large scale data analysis for environmental narratives and framing. Our work is largely inspired by Holder et al. \citep{holder-etal-2023} and Rowland et al., \citep{rowlands-etal-2024-predicting} who released work similar to ours, but focused on natural language processing and textual data in O\&G marketing ads.

Lastly, related work on video datasets mostly emphasizes benchmarking and assessing model capabilities \citep[e.g.,][]{patraucean2023perception, chandrasegaran2024hourvideo, chen2024rextime}. While we highly appreciate such work, this is only partially the focus of our paper. More importantly, we view our dataset artifact  as a tool to evaluate and improve video processing models. This in turn can facilitate computational narrative and framing analysis, and can provide computational assistance for large-scale social science studies, and finally to measure O\&G framing over time.

\subsection{Dataset Details}

\subsubsection{Detail of \textsc{Facebook}} \label{appendix:dataset_detail_facebook}
We stored the dataset in a JSON Lines file, where each line corresponds to a video, including not only labels but also metadata such as ID, URL, entity name, and video length in seconds.
Video resolution varies (e.g., 398x224, 400x400) based on the original content, and all videos are in MP4 format, including audio (though some videos lack original audio tracks).
The following is an example object of each line of the file:
{
\footnotesize
\begin{lstlisting}
{
  "video_id": "video_001",
  "video_url": "[ANONYMIZED]",
  "labels": ["PA"],
  "video_length_seconds": 15,
  "entity_name": "[ANONYMIZED]"
}
\end{lstlisting}
}
Note that anonymization is applied only in this paper; the actual released dataset maintains the original information.

\subsubsection{Evaluation of Distant Labels of the \textsc{Facebook} Domain} \label{appendix:facebook_distant_label_aval}

In the \textsc{Facebook} domain, there may be cases where video content and labels do not perfectly align, because the original labels were created mainly for textual content.
To evaluate this, one of the authors manually annotated 20 randomly selected videos based only on videos, comparing them with the original labels.
F-score between the original annotations and our manual video-based annotations was 83\%, suggesting that the distant labels are still reasonable quality.

\subsubsection{Detail of \textsc{YouTube}} \label{appendix:dataset_detail_youtube}

We stored the dataset in a JSON Lines file, with each line containing metadata including video ID, URL, published date, entity name, video length in seconds, entity's headquarter country and region, channel name and ID, and video view count, in addition to annotated labels.
The following is an example object of each line of the file:
{
\footnotesize
\begin{lstlisting}
{
  "video_id": "video_101",
  "video_url": "[ANONYMIZED]",
  "labels": ["Economy and Business", "Work", 
             "Environment", "Green Innovation"],
  "video_publish_date": "2021-11-11",
  "video_title": "[ANONYMIZED]",
  "video_length_seconds": 79,
  "video_views": 12000,
  "entity_name": "[ANONYMIZED]",
  "entity_country": "[ANONYMIZED]",
  "entity_region": "[ANONYMIZED]",
  "channel_id": "[ANONYMIZED]",
  "channel_name": "[ANONYMIZED]"
}
\end{lstlisting}
}
Note that anonymization is applied only in this paper; the actual released dataset maintains the original information.

\subsubsection{Construction Detail of \textsc{YouTube}}\label{appendix:dataset_construction_detail_of_youtube}

As described before, we initially retrieved up to 30 videos per company by searching ads. 
The number of 30 results from the restriction of the search results by a Python library we used.
We use `advertisment' as a searching query, where it contains unintentional spelling mistakes.
We, however, found many of the results by the query and the correct query `advertisement'' are identical.
Although channel-search introduces potential biases, our priority was collecting advertisement-like videos. 

To apply Fleiss' kappa to the multi-label task, the inter-annotator agreement was calculated by counting whether a label exists for each label, and the final score was computed. 
In the initial round of the guideline refining process, three annotators achieved an agreement score of 0.35 on 16 randomly selected videos. After guideline revisions, two annotators reached a 0.59 agreement score on a new set of randomly selected videos in the second round. Following further guideline improvements, the final round achieved a 0.61 agreement score between two annotators on 21 additional videos. 
On the other hand, calculating the kappa for each label and taking the average yielded 0.46. 
The `Patriotism' label was excluded from the calculation because it lacked a label. Additionally, the `Work' label had a kappa of -0.05, which appears to stem from an annotator's careless mistake. Although the calculation of inter-annotator agreement is subject to such constraints, GPT-4.1's F-score exceeds 70\% on average, suggesting that relatively reliable annotations have been performed.

These two annotators then completed the annotation for all 386 videos using this finalized guideline.
Note that, we initially attempted to simultaneously annotate whether framing was explicit or implicit. However, this was not adopted in the final annotation due to a lack of agreement levels.

\subsection{Annotation Examples}

Here we show some annotation examples in our guideline for the \textsc{YouTube} domain. Note that the following annotation examples are based on the annotator's impressions and do not represent any particular position.

\noindent
\textbf{Video URL: \url{https://www.youtube.com/watch?v=xRXA9xR_8o8}.}
This video gives the viewer the impression that energy is necessary to support industry and everyday life. It also includes an environmentally friendly image with a narrative ``minimal environmental impact''. It also includes images of workers.
Finally, the labels can be `Economy and Business', `Work', `Environment', and `Community and Life'.

\noindent
\textbf{Video URL: \url{https://www.youtube.com/watch?v=mVRC08LXfg0}.}
This video gives the viewer the impression that the company supports charitable efforts (Community and Life). As an implicit message, the company is supporting climate actions.
Finally, the labels can be `Community and Life' and `Environment'.

\subsubsection{Further Analysis}

The label distribution can be found in \Cref{tb:label_distribution}.

\begin{table}[t]
\centering
\footnotesize
\caption{The label distribution} \label{tb:label_distribution}
\begin{tabular}{lrr}
\toprule
 &  \textsc{YouTube} & \textsc{Facebook} \\
\midrule
\textbf{Label (Train/Test)} &
\begin{tabular}{r} Comm.\&Life (125/122)\\Work (102/96)\\Env. (84/78)\\Econ.\&Bus. (53/41)\\Green Innov. (39/28)\\Patriotism (19/15) \end{tabular} &
\begin{tabular}{r} CA (27/23)\\CB (33/29)\\PA (55/64)\\PB (10/7)\\GA (15/21)\\GC (30/33)\\SA (21/13) \end{tabular} \\
\bottomrule
\end{tabular}
\end{table}

\begin{figure}[t]
\centering
  \includegraphics[width=.4\linewidth]{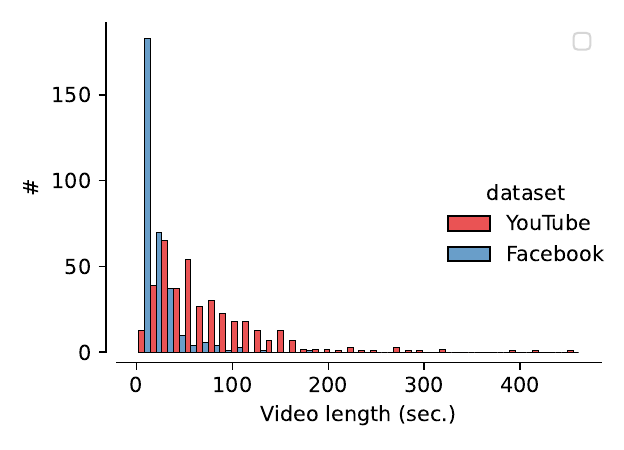}
  \caption{The video length distribution.}\label{fig:length_distribution}
\end{figure}

The video length distribution can be found in \Cref{fig:length_distribution}.

\begin{figure}[t]
    \centering
    \includegraphics[width=.8\linewidth]{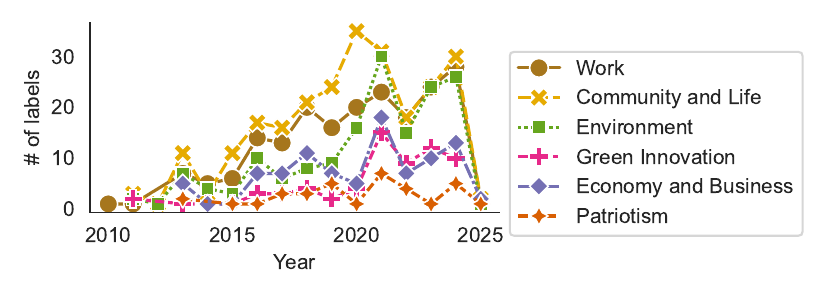} 
    \caption{The temporal distribution of \textsc{YouTube}. } \label{fig:yt_temporal_statistics}
\end{figure}

The temporal label distribution can be found in \Cref{fig:yt_temporal_statistics}.

\newpage

\subsubsection{Full Entity List} \label{appendix:full_entity_list}

\Cref{tb:entity_facebook} and \Cref{tb:entity_youtube} shows the entity list of the dataset from \textsc{Facebook} and \textsc{YouTube}, respectively.

\begin{table}[t]
\caption{Entity list of \textsc{Facebook}} \label{tb:entity_facebook}
\centering
\scriptsize
\begin{longtable}{lr}
Name & \# Video  \\
\hline
AGA & 35 \\
API & 50 \\
Alliancefor\_MI & 4 \\
BP & 7 \\
CA\_forAffordableandReliableEnergy & 1 \\
CA\_forEnergyIndependence & 29 \\
CO\_forResponsibleEnergyDevelopment & 2 \\
ConsumerEnergy & 4 \\
Enbridge & 4 \\
EnergyTransfer & 40 \\
ExxonMobil & 33 \\
GreatLakes\_MI & 4 \\
NMOGA & 48 \\
OH\_Oil\&Gas & 5 \\
Partnership\_EnergyProgress & 2 \\
TX\_Oil\&Gas & 3 \\
Williams & 49 \\
\\
\end{longtable} 
\end{table}

\begin{table}[t]
\caption{Entity list of \textsc{YouTube}} \label{tb:entity_youtube}
\centering
\scriptsize
\begin{tabular}{lrrr}
Name & Country by headquarters & Region & \# Video \\
\hline
Abu Dhabi National Oil Company & AE & Middle East & 20 \\
Alinta Energy & AU & Oceania & 13 \\
Ampol Limited & AU & Oceania & 9 \\
BP plc & GB & Europe & 14 \\
Bharat Petroleum Corporation Limited & IN & Asia & 8 \\
Cenovus Energy Inc. & CA & North America & 1 \\
Chevron Corporation & US & North America & 14 \\
China National Offshore Oil Corporation & CN & Asia & 2 \\
China Petroleum \& Chemical Corporation (Sinopec) & CN & Asia & 20 \\
ConocoPhillips & US & North America & 9 \\
Coterra Energy Inc. & US & North America & 8 \\
ENEOS Corporation & JP & Asia & 6 \\
Ecopetrol S.A. & CO & South America & 2 \\
Enbridge Inc. & CA & North America & 24 \\
Equinor ASA & NO & Europe & 4 \\
Exxon Mobil Corporation & US & North America & 12 \\
Galp Energia SGPS, S.A. & PT & Europe & 2 \\
Halliburton Company & US & North America & 5 \\
INPEX Corporation & JP & Asia & 2 \\
Indian Oil Corporation & IN & Asia & 6 \\
Kinder Morgan, Inc. & US & North America & 1 \\
Marathon Petroleum Corporation & US & North America & 3 \\
Neste Oyj & FI & Europe & 20 \\
OMV Aktiengesellschaft & AT & Europe & 19 \\
Occidental Petroleum Corporation & US & North America & 3 \\
Oil and Natural Gas Corporation & IN & Asia & 2 \\
Origin Energy Limited & AU & Oceania & 14 \\
PKN Orlen & PL & Europe & 3 \\
PTT Public Company Limited & TH & Asia & 6 \\
Phillips 66 Company & US & North America & 6 \\
Pioneer Natural Resources Company & US & North America & 1 \\
Repsol S.A. & ES & Europe & 21 \\
S-OIL Corporation & KR & Asia & 1 \\
SLB (Schlumberger Limited) & US & North America & 5 \\
SNAM S.p.A. & IT & Europe & 7 \\
Saudi Aramco & SA & Middle East & 23 \\
Shell plc & GB & Europe & 18 \\
Suncor Energy Inc. & CA & North America & 1 \\
TC Energy Corporation & CA & North America & 12 \\
TotalEnergies SE & FR & Europe & 18 \\
Valero Energy Corporation & US & North America & 16 \\
Woodside Energy Group Ltd & AU & Oceania & 5 \\
\\
\end{tabular} 
\end{table}

\begin{figure}[t]
\centering
  \includegraphics[width=.5\linewidth]{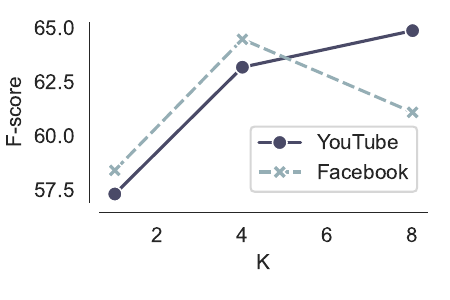}
  \caption{The effect of few-shot size for Qwen2.5-VL 7B.}\label{fig:fewshot}
\end{figure}

\subsection{Effect of Few-shot Size} \label{appendix:fewshot}

We investigate the effect of increasing the few-shot size ($K$) using Qwen2.5-VL 7B.
We set $K=1, 4, 8$ and evaluate the F-scores for both \textsc{YouTube} and \textsc{Facebook}.
To secure enough search space for selecting few-shot samples, we did not use ER in this experiment.
\Cref{fig:fewshot} shows the general trend that increasing $K$ improves F-scores.
Increasing the few-shot size takes longer inference time and higher computational cost, which is the limitation resulted from the trade-off between performance and efficiency.

\subsection{Implementation Detail} \label{appendix:implementation}

We use PyTorch 2.6.0 \cite{pytorch}, HuggingFace Transformers 4.51.2 \cite{wolf-etal-2020-transformers}, and vLLM 0.8.3 \cite{vllm} libraries for the model and inference implementation. 
For GPT-4.1, we use the version of `gpt-4.1-2025-04-14'.
For GPT-4o-mini, we use the version of `gpt-4o-mini-2024-07-18'.
For Qwen2.5-VL, we use the versions of `Qwen/Qwen2.5-VL-7B-Instruct' and `Qwen/Qwen2.5-VL-32B-Instruct'.
For InternVL2, we use the version of `OpenGVLab/InternVL2-8B'.
For DeepSeekVL2, we use the version of `deepseek-ai/deepseek-vl2'.
For the CLIP embedding, we use `openai/clip-vit-base-patch32' available on Huggingface.
This study does not conduct any hyperparameter search and model selection.
For all VLMs, we set temperature at 0.
Due to limited computational resources of us, we only conduct the single run and report the F-score for each VLM.

\begin{figure*}[tp]
    \centering
    \includegraphics[width=\linewidth]{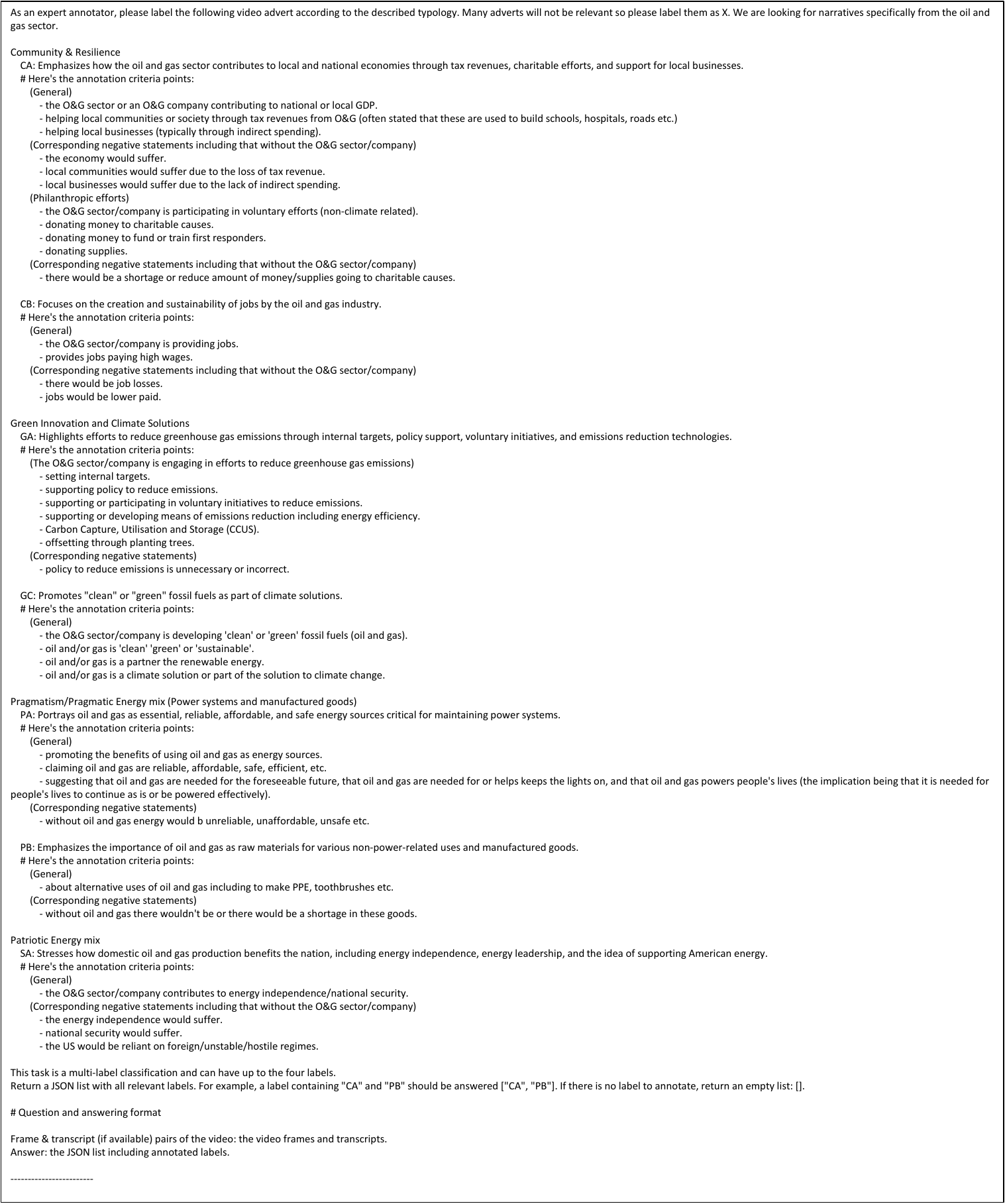}
    \caption{The guideline part of the prompt for \textsc{Facebook}.} \label{fig:prompt_fb}
\end{figure*}

\begin{figure*}[tp]
    \centering
    \includegraphics[width=\linewidth]{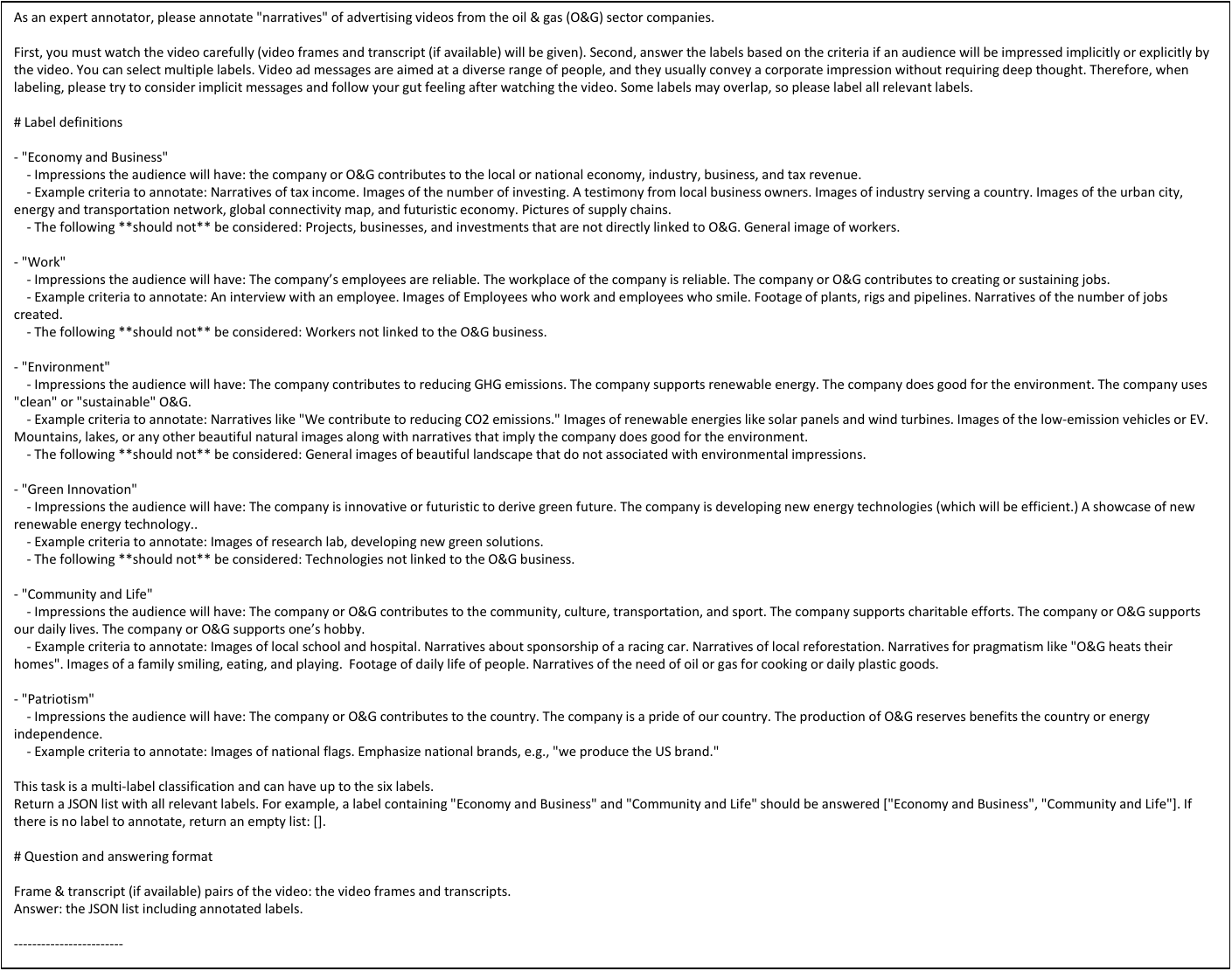}
    \caption{The guideline part of the prompt for \textsc{YouTube}.} \label{fig:prompt_yt}
\end{figure*}

\Cref{fig:prompt_fb} and \Cref{fig:prompt_yt} shows the beginning of the prompt (i.e., the annotation guideline part) of \textsc{Facebook} and \textsc{YouTube}, respectively.

\paragraph{Detail of Image Frames Extraction}
Transcript segments with very short text (three words or fewer) are excluded, as we found they tend to be noisy.
If the number of segments exceeds $N_{\text{Frame}}$, excess segments and their corresponding frames are discarded.
If no transcript is available (i.e., $N_{\text{Frame}} = 0$), we sample $N_{\text{Frame}}$ frames uniformly from the video. In this case, the corresponding transcript segments are set as ``N/A''.

\newpage

\subsection{Computational Resources} \label{appendix:computational_resource}

We use Amazon EC2's `g5.48xlarge' (8 x  A10G Tensor Core GPU (192 GiB GPU memory) and 768 GiB memory) for experiments using open-source VLMs.
The experiment was conducted over a period of about two days.

\subsection{Full Results of the Ablation Study} \label{appendix:ablation_full}

\Cref{tb:ablation_full} shows the full results of the ablation study.

\begin{table}[t]
\centering
\footnotesize
\caption{Ablation results in the 1-shot prediction. T represents the transcript input, ES represents the embedding search in the 1-shot sample selection, and ER represents the entity restriction in the 1-shot sample selection.} \label{tb:ablation_full}
\begin{tabular}{lllllrr}
\toprule

\textbf{Model} & \textbf{Params} & \textbf{T}  &  \textbf{ES} & \textbf{ER}  & \textsc{YouTube} & \textsc{Facebook} \\
\midrule
\rowcolor[rgb]{0.93,0.93,0.93}DeepSeekVL2 & 4.5B & $\checkmark$ & $\checkmark$ & $\checkmark$ & 49.7 & 62.3 \\
 &  & $\times$ & $\checkmark$ & $\checkmark$ & \textbf{53.5} & \textbf{63.8} \\
 &  & $\checkmark$ & $\times$ & $\checkmark$ & 46.9 & 43.8 \\
 &  & $\checkmark$ & $\checkmark$ & $\times$ & 50.5 & 60.6 \\
\midrule
\rowcolor[rgb]{0.93,0.93,0.93}InternVL2 & 8B & $\checkmark$ & $\checkmark$ & $\checkmark$ & 56.7 & 46.2 \\
 &  & $\times$ & $\checkmark$ & $\checkmark$ & \textbf{58.3} & \textbf{48.4} \\
 &  & $\checkmark$ & $\times$ & $\checkmark$ & 52.1 & 39.6 \\
 &  & $\checkmark$ & $\checkmark$ & $\times$ & 56.3 & 46.1 \\
\midrule
\rowcolor[rgb]{0.93,0.93,0.93}Qwen2.5-VL & 7B & $\checkmark$ & $\checkmark$ & $\checkmark$ & 59.2 & 58.2 \\
 &  & $\times$ & $\checkmark$ & $\checkmark$ & \textbf{60.0} & 53.6 \\
 &  & $\checkmark$ & $\times$ & $\checkmark$ & 54.3 & 48.2 \\
 &  & $\checkmark$ & $\checkmark$ & $\times$ & 57.3 & \textbf{58.4} \\
\midrule
\rowcolor[rgb]{0.93,0.93,0.93}Qwen2.5-VL & 32B & $\checkmark$ & $\checkmark$ & $\checkmark$ & \textbf{66.2} & \textbf{70.5} \\
 &  & $\times$ & $\checkmark$ & $\checkmark$ & 61.2 & 60.6 \\
 &  & $\checkmark$ & $\times$ & $\checkmark$ & 64.0 & 59.1 \\
 &  & $\checkmark$ & $\checkmark$ & $\times$ & 65.6 & 68.1 \\
\bottomrule
\end{tabular}
\end{table}

\subsection{Declaration of LLM Usage} \label{appendix:declaration_of_LLM_usage}

We used ChatGPT and Gemini 2.5 in parts of our paper to translate, correct grammar, and polish the writing.

\newpage

\subsection{Dataset Documentation} \label{appendix:dataset_documentation}

Based on the dataset sheets by Gebru et al. \cite{gebru-etal-2021-datasheets}, the following provides information of our dataset.

\subsubsection{Motivation}

\paragraph{For what purpose was the dataset created?}
We created this annotated video dataset to benchmark VLMs for predicting obstruction and impressionistic framing by O\&G entities. 

\paragraph{Who created the dataset (for example, which team, research group) and on behalf of which entity (for example, company, institution, organization)?}
A research team of the authors from Stanford University, Princeton University, InfluenceMap, and Hitachi America.

\paragraph{Who funded the creation of the dataset?}
No entity explicitly funded the creation of the dataset, but GM conducted this study within the Stanford Data Science (SDS) Affiliates Program.

\subsubsection{Composition}

\paragraph{What do the instances that comprise the dataset represent (for example, documents, photos, people, countries)?}
The dataset includes videos, their metadata and annotated labels.
We cover video publishers from different countries (at headquarters-level).

\paragraph{How many instances are there in total (of each type, if appropriate)?}
See \Cref{tb:basic_statistics} and \Cref{tb:label_distribution}.

\paragraph{Does the dataset contain all possible instances or is it a sample (not necessarily random) of instances from a larger set?}
The video collection process includes sampling. In particular, videos were collected from the \textsc{YouTube} domain by searching official channels, which involves a black box process. For details, see \Cref{appendix:dataset_construction_detail_of_youtube}.

\paragraph{What data does each instance consist of?}
Our dataset mainly consists of video metadata (such as URL) and annotated labels. See \Cref{appendix:dataset_detail_facebook} and \Cref{appendix:dataset_detail_youtube} for more detail.

\paragraph{Is there a label or target associated with each instance?}
Yes, the framing labels are associated with each video instance.

\paragraph{Is any information missing from individual instances?}
N/A.

\paragraph{Are relationships between individual instances made explicit (for example, users’ movie ratings, social network links)?}
N/A.

\paragraph{Are there recommended data splits (for example, training, development/validation, testing)?}
Yes. There is an official data split for training and testing.

\paragraph{Are there any errors, sources of noise, or redundancies in the dataset?}
The annotations can contain errors by human mistakes or unrecognized biases.
Some videos can be similar, e.g., an entity publishes similar videos based on a series content.

\paragraph{Is the dataset self-contained, or does it link to or otherwise rely on external resources (for example, websites, tweets, other datasets)?}
The video content can be accessible by a link.

\paragraph{Does the dataset contain data that might be considered confidential (for example, data that is protected by legal privilege or by doctor–patient confidentiality, data that includes the content of individuals’ non-public communications)?}
No.

\paragraph{Does the dataset contain data that, if viewed directly, might be offensive, insulting, threatening, or might otherwise cause anxiety?}
No.

\paragraph{Does the dataset identify any subpopulations (for example, by age, gender)?}
No.

\paragraph{Is it possible to identify individuals (that is, one or more natural persons), either directly or indirectly (that is, in combination with other data) from the dataset?}
No.

\paragraph{Does the dataset contain data that might be considered sensitive in any way (for example, data that reveals race or ethnic origins, sexual orientations, religious beliefs, political opinions or union memberships, or locations; financial or health data; biometric or genetic data; forms of government identification, such as social security numbers; criminal history)?}
No.

\subsubsection{Collection process}

\paragraph{How was the data associated with each instance acquired? Was the data directly observable (for example, raw text, movie ratings), reported by subjects (for example, survey responses), or indirectly inferred/ derived from other data (for example, part-of-speech tags, model-based guesses for age or language)?}
The annotated labels for videos were originally annotated by expert annotators.
Each video was directly observable to the annotators.
For the \textsc{Facebook} domain, it contains distant labels where we map the annotated labels on textual content to the video content automatically.
See \Cref{sec:dataset} for more detail.

\paragraph{What mechanisms or procedures were used to collect the data (for example, hardware apparatuses or sensors, manual human curation, software programs, software APIs)?}
See \Cref{sec:dataset}.

\paragraph{If the dataset is a sample from a larger set, what was the sampling strategy (for example, deterministic, probabilistic with specific sampling probabilities)?}
To extract target videos for annotations of \textsc{YouTube}, we sampled videos from the pool of available videos. The random sampling was applied for this process.

\paragraph{Who was involved in the data collection process (for example, students, crowdworkers, contractors) and how were they compensated (for example, how much were crowdworkers paid)?}
For \textsc{YouTube}, the authors engaged in the annotations. There are no compensations.

\paragraph{Over what timeframe was the data collected?}
See \Cref{sec:dataset} and \Cref{sec:dataset_analysis}.

\paragraph{Were any ethical review processes conducted (for example, by an institutional review board)?}
No.

\paragraph{Did you collect the data from the individuals in question directly, or obtain it via third parties or other sources (for example, websites)?}
We obtained video data from websites.

\paragraph{Were the individuals in question notified about the data collection?}
N/A.

\paragraph{Did the individuals in question consent to the collection and use of their data?}
N/A.

\paragraph{If consent was obtained, were the consenting individuals provided with a mechanism to revoke their consent in the future or for certain uses?}
N/A.

\paragraph{Has an analysis of the potential impact of the dataset and its use on data subjects (for example, a data protection impact analysis) been conducted?}
N/A.

\subsubsection{Preprocessing/cleaning/labeling}

\paragraph{Was any preprocessing/cleaning/labeling of the data done (for example, discretization or bucketing, tokenization, part-of-speech tagging, SIFT feature extraction, removal of instances, processing of missing values)?}
The labeling was conducted by annotators.
We transcribe the videos using Whisper-1.

\paragraph{Was the ``raw'' data saved in addition to the preprocessed/cleaned/ labeled data (for example, to support unanticipated future uses)?}
Yes.

\paragraph{Is the software that was used to preprocess/clean/label the data available?}
No.

\subsubsection{Uses}

\paragraph{Has the dataset been used for any tasks already?}
No.

\paragraph{Is there a repository that links to any or all papers or systems that use the dataset?}
No.

\paragraph{What (other) tasks could the dataset be used for?}
The dataset might be used for potential task of corporate climate engagement assessment and greenwashing detection.

\paragraph{Is there anything about the composition of the dataset or the way it was collected and preprocessed/ cleaned/labeled that might impact future uses?}
See \Cref{appendix:limitation}.

\paragraph{Are there tasks for which the dataset should not be used?}
This dataset should not be used to attack or mislabel real entities, locations, or individuals.

\subsubsection{Distribution}

\paragraph{Will the dataset be distributed to third parties outside of the entity (for example, company, institution, organization) on behalf of which the dataset was created?}
Yes.

\paragraph{How will the dataset be distributed (for example, tarball on website, API, GitHub)?}
See \Cref{appendix:code_and_data}.

\paragraph{When will the dataset be distributed?}
Before the camera ready deadline of this paper.

\paragraph{Will the dataset be distributed under a copyright or other intellectual property (IP) license, and/or under applicable terms of use (ToU)?}
Except for third-party content, we will license the dataset with CC BY-NC 4.0.

\paragraph{Have any third parties imposed IP-based or other restrictions on the data associated with the instances?}
Each video content is copyrighted by its respective publisher.

\paragraph{Do any export controls or other regulatory restrictions apply to the dataset or to individual instances?}
No.

\subsubsection{Maintenance}

\paragraph{Who will be supporting/hosting/maintaining the dataset?}
The authors.

\paragraph{How can the owner/curator/ manager of the dataset be contacted (for example, email address)?}
By the email address.

\paragraph{Is there an erratum?}
N/A.

\paragraph{Will the dataset be updated (for example, to correct labeling errors, add new instances, delete instances)?}
Yes. We may delete instances that associate with deleted videos.

\paragraph{If the dataset relates to people, are there applicable limits on the retention of the data associated with the instances (for example, were the individuals in question told that their data would be retained for a fixed period of time and then deleted)?}
N/A.

\paragraph{Will older versions of the dataset continue to be supported/hosted/ maintained?}
No.

\paragraph{If others want to extend/augment/build on/contribute to the dataset, is there a mechanism for them to do so? }
No.

\end{document}